\documentclass[conference]{IEEEtran}
\IEEEoverridecommandlockouts
\usepackage{cite}
\usepackage{amsmath}
\usepackage{amssymb}
\usepackage{algorithm}
\usepackage{bbm}
\usepackage{algorithmic}
\usepackage{graphicx}
\usepackage{textcomp}
\usepackage{xcolor}
\usepackage{verbatim}
\usepackage[skip=0pt,font=footnotesize]{caption}
\def\BibTeX{{\rm B\kern-.05em{\sc i\kern-.025em b}\kern-.08em
    T\kern-.1667em\lower.7ex\hbox{E}\kern-.125emX}}

\usepackage{times}
\usepackage{epsfig}
\usepackage{bbm}
\usepackage{booktabs}
\usepackage{algorithm}
\usepackage{algorithmic}
\usepackage [english]{babel}
\usepackage [autostyle, english = american]{csquotes}
\MakeOuterQuote{"}
\DeclareMathOperator*{\argmin}{argmin}

\newlength\myindent
\newcommand\bindent{%
	\begingroup
	\setlength{\itemindent}{\myindent}
	\addtolength{\algorithmicindent}{\myindent}
}
\newcommand\eindent{\endgroup}

\begin{document}

\title{Generalized Adversarial Distances to Efficiently Discover Classifier Errors}

\author{
\IEEEauthorblockN{Walter Bennette}
\IEEEauthorblockA{\textit{Information Directorate} \\
	\textit{Air Force Research Lab}\\
	Rome, NY \\
	walter.bennette.1@us.af.mil}
\and
\IEEEauthorblockN{Sally Dufek}
\IEEEauthorblockA{\textit{Department of Statistics} \\
	\textit{Miami University}\\
	Oxford, OH \\
	dufeks@miamioh.edu}
\and
\IEEEauthorblockN{Karsten Maurer}
\IEEEauthorblockA{\textit{Data Science} \\
	\textit{General Mills}\\
	Minneapolis, MN\\
	karsten.datasci@gmail.com}
\and
\IEEEauthorblockN{Sean Sisti}
\IEEEauthorblockA{\textit{Information Directorate} \\
	\textit{Air Force Research Lab}\\
	Rome, NY \\
	sean.sisti@us.af.mil}

\and
\IEEEauthorblockN{Bunyod Tusmatov}
\IEEEauthorblockA{\textit{Department of Statistics} \\
	\textit{Miami University}\\
	Oxford, OH \\
	tusmatbs@miamioh.edu}
}

\maketitle

\begin{abstract}

Given a black-box classification model and an unlabeled evaluation dataset from some application domain, efficient strategies need to be developed to evaluate the model.  Random sampling allows a user to estimate metrics like accuracy, precision, and recall, but may not provide insight to high-confidence errors.  High-confidence errors are rare events for which the model is highly confident in its prediction, but is wrong.  Such errors can represent costly mistakes and should be explicitly searched for.  In this paper we propose a generalization to the Adversarial Distance search that leverages concepts from adversarial machine learning to identify predictions for which a classifier may be overly confident.  These predictions are useful instances to sample when looking for high-confidence errors because they are prone to a higher rate of error than expected.  Our generalization allows Adversarial Distance to be applied to any classifier or data domain.  Experimental results show that the generalized method finds errors at rates greater than expected given the confidence of the sampled predictions, and outperforms competing methods.

\end{abstract}

%%%%%%%%% BODY TEXT
\section{Introduction}

When using a pre-trained black-box classification model, made available by a third party, users typically have little knowledge of how the model was trained.  As a result, the user may be unaware of how the model will perform in their specific environment. Therefore, the model should be thoroughly tested before it is deployed for some application.  For example, if a black-box model were to be used to detect fraudulent banking activity, it is necessary to evaluate the tool before it is utilized.  The evaluation procedure should reveal the effectiveness of the model for the specific application domain, and be respectful of the user's time and effort.

Many strategies can be developed to evaluate a classification model, but most assume the availability of a labeled test dataset.  However, such a test dataset is not always available, especially when a user is already depending on a pre-trained classification model.  Therefore, it is necessary to develop human-in-the-loop evaluation strategies for unlabeled evaluation datasets.  These strategies should limit the amount of human interaction and provide a useful assessment of the model.  

One evaluation strategy is to randomly sample instances from an unlabeled evaluation dataset, and have the user provide labels for the sampled instances.  This allows estimates to be calculated for measures like accuracy, precision, and recall.  Another strategy is to use the classification model to provide predictions for the entire evaluation dataset, and have the user inspect those predictions for which the classifier has low confidence.  This provides insight to areas where the model is unsure of its predictions.  Unfortunately, both of these strategies are likely to miss high-confidence errors.  High-confidence errors are predictions for which the model is highly confident in its prediction, but is wrong.  Such errors can have large consequences (e.g allowing fraudulent activity) and should be specifically searched for.

Causes of high-confidence errors can range from dataset bias during training \cite{Stock2017}, domain shift during use \cite{Sugiyama2017}, lack of model expressiveness, overfitting, and more.  Generally, high-confidence errors can be viewed as blind spots within classification models \cite{Attenberg2015}.  For example, \cite{Lakkaraju2016} presented a classification problem where the goal was to distinguish images of cats and dogs.  However, the training data was made exclusively of cats with light fur and dogs with dark fur.  When testing on an unbiased dataset, dogs with light fur were predicted to be cats with high confidence.  Finding this type of high-confidence error would enable the user to discover more about their classifiers' strengths and weaknesses.

Initial strategies for finding high-confidence errors in unlabeled datasets focused on discovering  errors above some confidence threshold, $\tau$ (generally set at $0.65$ for binary classification) \cite{Bansal2018, Lakkaraju2016}.  However, it was observed that this practice ignores the expectation that some amount of error is expected for confidence levels below 100\%.  Meaning, the methods could be discovering errors accidentally, not by discovering commonalities between errors to increase the rate of error discovery.  More recent works have instead focused on discovering classification errors at rates greater than expected, to encourage search methods that discover something about a model's weaknesses to increase the rate of error discovery \cite{Maurer2018, Bennette2020}.  Of particular interest to this work is a search technique that leverages adversarial machine learning to point users towards high confidence errors.  Although the method proves to be very effective at finding the errors of interest, the method is only defined for image classification models \cite{Bennette2020}.  

In this work we introduce a general approach to finding errors through adversarial machine learning.  Specifically, we adapt the Adversarial Distance Search \cite{Bennette2020} to be applicable to any data modality and classification algorithm.  In our approach we develop a general adversarial black-box attack which is used to help find high confidence errors. This is accomplished by comparing the  amount of perturbation required to change the classifier's prediction to the amount of perturbation expected, given the confidence of the prediction.  We believe predictions that require less perturbation than expected are good instances to search for high-confidence errors.

The remainder of this paper is organized as follows: in Section \ref{sec:RelatedWork} we discuss related works, in \ref{sec:ProblemFormulation} we introduce our problem formulation, in \ref{sec:Methodology} we present our methodology, in \ref{sec:Results} we discuss our experimental results, and in \ref{sec:Conclusions} we conclude while providing thoughts for future research.

\section{Related Work}\label{sec:RelatedWork}

 In this section we introduce existing human-in-the-loop sampling strategies for finding high-confidence errors, discuss adversarial machine learning as it pertains to our approach, and briefly discuss model extraction techniques.
 
\subsection{High-Confidence Errors} 

Algorithmic approaches for finding high-confidence errors from machine learning classifiers mainly consist of the following components: 1) a utility function to measure the usefulness of queried points, 2)  some search strategy to help maximize the utility function \cite{ Bansal2018, Lakkaraju2016, Maurer2018}

In Lakkaraju (2017) \cite{Lakkaraju2016}, they defined a utility function that gave a uniform value for each error discovered above some threshold (65\% for binary classification), and applied a penalty associated with the cost of the human providing the true label for the queried points.  The suggested search was then to cluster data points from the unlabeled evaluation dataset using some feature space (which, in the case of black-box classifiers, may be different than the original feature space), and perform multi-armed bandit sampling to maximize the utility.  

Bansal and Weld (2018) \cite{Bansal2018} defined a utility function that encouraged the discovery of high-confidence errors dispersed throughout some features space.  The intent of encouraging the discovery of diverse errors was to prevent the query algorithm from repeatedly sampling a rich pocket of errors.  The suggested search technique was then to cluster the feature space of the evaluation data, and perform a greedy selection over the clusters to maximize the utility.  Of special note is that \cite{Bennette2020} experimentally showed that the search techniques in \cite{Lakkaraju2016, Bansal2018, Bennette2020}, and random sampling all query similarly spread out instances.  Meaning, the addition of a diversity reward to the utility function may not be of practical concern.  

Maurer and Bennette (2019) \cite{Maurer2018} identified an oversight in the previous works.  That is, a certain amount of error should be expected for predictions made with anything less than 100\% confidence.  By defining a high-confidence error to simply be any error over some threshold, the previous techniques may be rewarding random error discovery.  Instead, \cite{Maurer2018} suggested a new quality measure that compares the actual number of discovered errors to the expected number of errors given the confidence of the model's predictions.  As such, discovering errors by chance would be rewarded less than discovering errors due to leveraging some knowledge of the classifier's underling error structure. This concept is captured in a measure coined the Standardized Discovery Ratio (SDR), which is discussed in greater detail in Section \ref{sec:ProblemFormulation}.  The utility function of \cite{Maurer2018} was then a combination of the SDR and a diversity award.  The suggested search technique relied on a logistic regression meta-model to try and learn the types of predictions prone to error.

Finally, Bennette (2020) \cite{Bennette2020} suggested an Adversarial Distance search, which relies on concepts from adversarial machine learning, to identify promising instances to query during a search for high-confidence errors.  For images, Adversarial Distance quantifies how much perturbation is required to change a targeted classifier's prediction for a specific image, compared to the expected amount of perturbation given the confidence of the classifier's prediction.  The intuition behind this measure comes from an observation that some images require more perturbation to change a targeted classifier's prediction than other images that receive a similar prediction and confidence score.  It is hypothesized that the targeted classifier is basing its prediction on non-robust features when an image requires less perturbation than expected.  Non-robust features are introduced in \cite{Ilyas2019} as highly discriminant features inherent to datasets, but not related to the human perceived classification task.  It is also observed in \cite{Ilyas2019} that these non-robust features are brittle and easily broken by adversarial attacks.  Therefore, when searching for high-confidence errors, Adversarial Distance can help identify images for which a classifier is overly confident; possibly due to the presence of non-robust features in the image.  In \cite{Bennette2020} the quality of sampled points was measured by SDR and the search technique was to query images with low Adversarial Distance.  In this work we want to generalize Adversarial Distance to other data modalities because of its promising performance when using the measure to drive searches for high confidence errors within the image domain.

\subsection{Adversarial Machine Learning}

Adversarial machine learning is a growing field of interest that attempts to attack machine learning solutions.  During inference, one purpose of an adversarial attack is to fool a classification model into making an incorrect prediction \cite{Goodfellow2015}.  Many attack methods have been developed and they primarily focus on computer vision tasks \cite{Goodfellow2015, Brendel2017, rauber2017}.  However, recent efforts have also introduced attacks for natural language processing tasks \cite{Gao2018nlp}, and for tabular data \cite{Ballet2019}.  Regardless of the modality of the data (imagery, text, or tabular), the goal of the attacks are to minimally change an instance in such a way that a human will not recognize a significant difference, but the classifier will change its prediction.  

Adversarial attacks used to alter input data can be grouped into white-box or black-box attacks.  White-box attacks assume access to the weights and architecture of the classifier being attacked.  Black-box attacks assume limited knowledge of the classifier being attacked, and typically only require the black-box to provide predictions.  In the Adversarial Distance search from \cite{Bennette2020} they rely on a black-box attack for image classification called the Boundary Attack \cite{Brendel2017}.  The purpose is to allow the evaluation of any image classifier.  However, this attack is exclusive to imagery, and will not satisfy our needs for generalizing Adversarial Distance.

When divorced from a specific data domain, an adversarial attack is simply an attempt to find the smallest amount of change required to push a data point across the decision boundary of a targeted classifier.  Unfortunately, if the classifier's decision boundaries are unknown, as would be the case with a black-box classifier, finding this distance is non-trivial.  One possible solution is to learn a pseudo model that behaves similarly to the targeted classifier and then use that model to help develop the attack \cite{Brendel2017, Ballet2019}.  This approach requires the extraction of a pseudo model which could be challenging with limited access to the black-box model, but has shown to be feasible when the black-box model can be sufficiently probed \cite{Bastani2017, Tramer2016}.  In Section \ref{subsec:GAA} we introduce a general adversarial attack that would be appropriate for a black-box classifier from any domain, and is suitable for generalizing Adversarial Distance.

\section{Problem Formulation}\label{sec:ProblemFormulation}

Early methods for finding high-confidence errors seemingly ignore the fact that prediction errors should occur at confidence levels below 100\% \cite{Bansal2018, Lakkaraju2016}.  By rewarding the discovery of errors above some confidence threshold, these search methods may simply be discovering errors by random chance, not from a deeper understanding of the error structure found within the classifier.  Recently, works such as \cite{Maurer2018} and \cite{Bennette2020} explicitly encourage the discovery of errors at rates that exceed expectations by measuring the Standardized Discovery Ratio (SDR). SDR can be viewed as the number of discovered misclassifications relative to what would be expected given the confidence of the predictions. This valuation rewards search methods that uncover model weaknesses to increase error discovery, rather than random chance.  SDR is defined more strictly below.

Here we present the problem formulation as introduced in \cite{Maurer2018} and \cite{Bennette2020}.  Given a black-box classifier, $M$, with $M(x) = (\hat{y}_x, \hat{p}_x)$, where $x$ is an instance from an unlabeled evaluation set $X$, $\hat{y}_x$ is the model's prediction,  $\hat{p}_x$ is the model's confidence of that prediction, and $y_x $ is the true label assigned by some oracle (human-in-the-loop), the task is to find a query set of data points, $Q \subseteq X$, that maximize the SDR.  SDR is defined as:

\begin{equation}
SDR = \frac{\sum_{q \in Q} \mathbbm{1}(\hat{y}_q \neq y_q)}{\sum_{q \in Q}\left(1-\hat{p}_q\right)},
\label{eq:sdr}
\end{equation}

\noindent where $|Q| = B$, $\hat{p}_q > 0.65$ for $q \in Q$, and $B$ represents the labeling budget of the oracle used to find true labels.  

As with the prior works, our goal is to devise a sampling strategy that maximizes SDR.

\section{Generalized Adversarial Distance Search}\label{sec:Methodology}

This section introduces a methodology to perform a human-in-the-loop search for high-confidence errors based on a generalization of Adversarial Distance.  The original work \cite{Bennette2020}, defined Adversarial Distance only for image classification, but here we extend the idea such that it is possible to work with additional classification models and data modalities.

\subsection{Generalized Adversarial Distance}

Recall the description of the Adversarial Distance search from Section \ref{sec:RelatedWork}.  To generalize Adversarial Distance we need to replace the black-box image classification attack in the original formulation with a black-box attack appropriate for any data domain. We introduce a possible solution to this problem further down in Section \ref{subsec:GAA}. Then, Generalized Adversarial Distance (GAD) can be defined similarly to that in \cite{Bennette2020}: 

\begin{equation}
GAD(x) = MAE\left(x, A(M,x) \right) - F\left(\hat{p}_x\right),
\label{eq:GAD}
\end{equation}

\noindent where $x$ is the instance for which we are calculating the GAD, $M$ is the classifier being evaluated, $A(M,x)$ is a mechanism to create an adversarial version of $x$ such that $M$'s prediction is changed, and $F\left(\hat{p}_x\right)$ is a function to calculate the expected Mean Absolute Error (MAE) between $x$ and its adversarial partner based on the classifier's predictive confidence, $\hat{p}_x$.

For two instances, $a$ and $b$, each with $M$ features, MAE is defined to be:

\begin{equation}
MAE(a,b) = \frac{1}{M} \sum_{i=1}^M |a_{(i)} - b_{(i)}|.
\label{eq:MAE}
\end{equation}

\begin{figure}[hbt!]
	\begin{center}
		\includegraphics[width=1\linewidth]{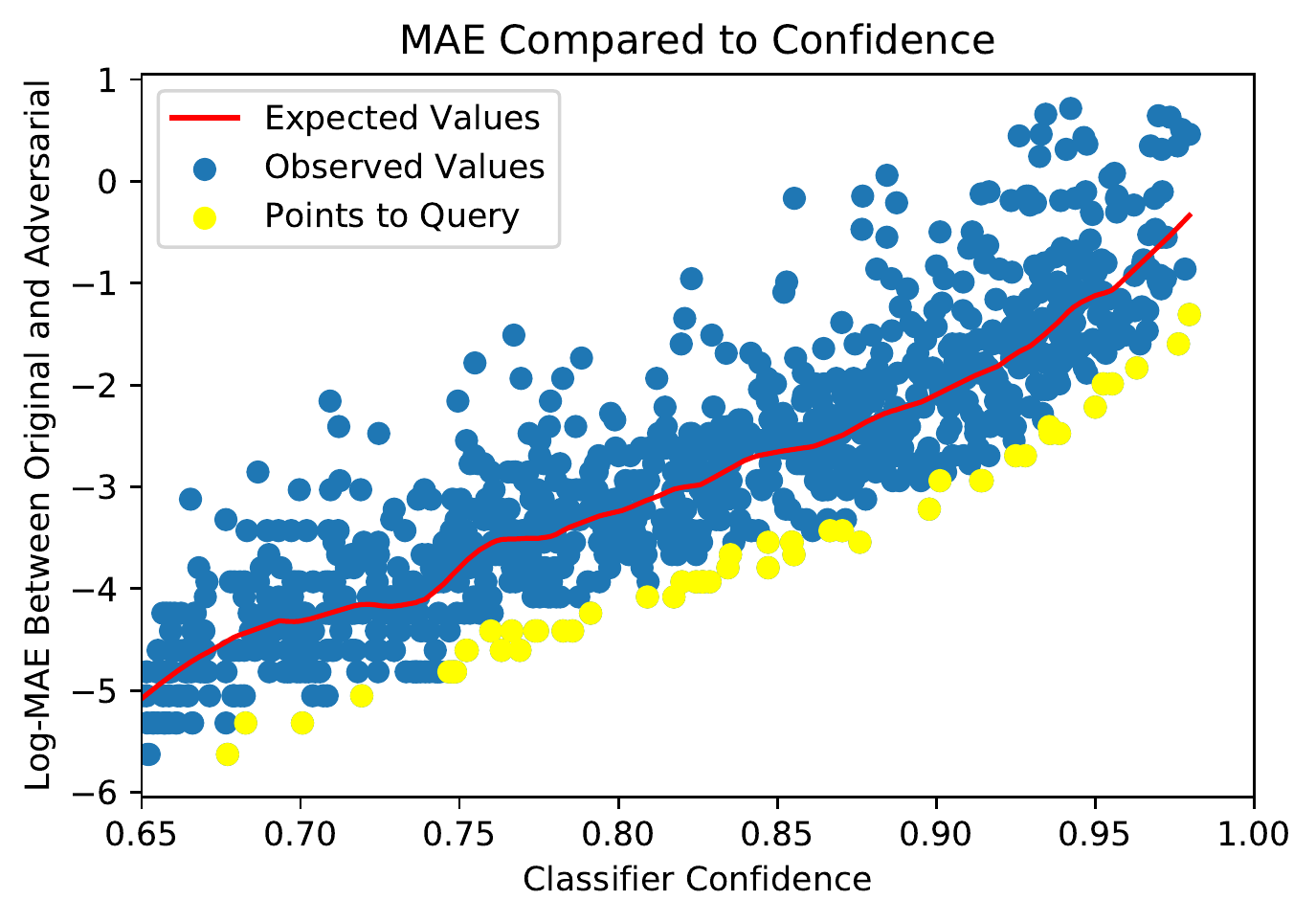}
	\end{center}
	\caption{Example LOESS curve fit between log-MAE and classifier confidence.  The vertical distance of points from the fitted line represents the Adversarial Distance, thus the yellow points are the farthest below the LOESS line and have the smallest Adversarial Distances. The yellow instances would be used to query the oracle and search for errors.}
	\label{fig:curve}
\end{figure}

$F$, the function used to estimate the MAE between an instance and its adversary for a given confidence, can be estimated by calculating the MAE between every instance in an evaluation set and its adversary, and then fitting a LOESS \cite{cleveland1988} regression line where the classifier confidence is the independent variable and MAE is the dependent variable (the log of MAE may be taken to facilitate better curve fitting).  Figure \ref{fig:curve} provides an example LOESS curve for the Pang04 dataset described later in the paper.  Observe that the vertical distance of the points from the fitted line represents the GAD.  Meaning, the points farthest below the LOESS line will have the smallest GAD.  Additionally, observe that GAD is completely unsupervised because the true labels of the instances are not used.

\subsection{Generalized Adversarial Attack}\label{subsec:GAA}

Our main contribution to generalizing Adversarial Distance is providing an adversarial attack, $A(M,x)$, to turn a data point adversarial for any classification model.  Our adversarial attack does require a pseudo model to be extracted from the targeted black-box model, but existing techniques for model extraction show that given a sufficient number of model queries, an accurate pseudo model can be obtained in most scenarios \cite{Bastani2017, Tramer2016}.  In this research we assume full access to the black box model, and there are no restrictions on the number of predictions that can be requested from the black-box classifier for extracting the desired pseudo model.  As a result, we are able to extract pseudo-models that prove useful when turning instances adversarial for our evaluation datasets.

\begin{algorithm}[hbt!]
	\caption{Model Extraction}
	\label{alg:extract}
	\begin{algorithmic}
		\STATE {\bfseries Input:} Original classifier $M_O$, representative data from application domain $D$ with $M$ features, size of design space $N$, an untrained Deep Neural Network $M_P$, number of training epochs $e$, and a specific class of interest.
		\STATE
		\STATE {\bfseries Step 1:} Create a Latin Hypercube Space, $L$, of $N$ x $M$ points such that the values found in each column of the design match the range of values found in the respective column of $D$. 
		\STATE {\bfseries Step 2:} Train $M_P$ for $e$ epochs to minimize the Mean Square Error between the confidence of the predictions for $M_O(L)$ and $M_P(L)$ when considering the specific class of interest
		\STATE {\bfseries Return:} $M_P$, a pseudo-model trained to predict the confidence $M_O$ will report for the class of interest for new instances
	\end{algorithmic}
	\label{alg:extract}
\end{algorithm}

Algorithm \ref{alg:extract} presents our model extraction paradigm.  This algorithm creates a pseudo model, $M_P$, that will estimate the original model's confidence for a specific class prediction.  This is achieved by having the original model, $M_O$, label a training dataset.  The training dataset is created by deploying a Latin Hybercube Sampling (LHS) design that matches the domain of an unlabeled evaluation datasest.  Then, $M_O$ provides prediction confidences for the instances generated by the LHS design, for a specific class of interest.  Finally, $M_P$ is some form of Deep Neural Network (DNN) which we train to minimize the Mean Square Error between the prediction confidences for the class of interest of $M_O$ and $M_P$, over the generated training data.  Note that if a sufficiently large unlabeled evaluation dataset can be obtained, it can be substituted for the LHS design.  This route may be preferable for complex domains such as image or text classification.

\begin{algorithm}[hbt!]
	\caption{Adversarial Attack}
	\label{alg:adv}
	\begin{algorithmic}
		\STATE {\bfseries Input:} Original classifier $M_O$, pseudo model $M_P$, instance to turn adversarial $x$, step size for search $\epsilon$
		\STATE
		\STATE $x_{adv} = x$, create a copy of $x$ to manipulate
		
		{\bfseries While:} $M_O(x) = M_O(x_{adv}) $ {\bfseries do:}
		\bindent
		\STATE $c = conf(M_P(x))$, confidence of $M_P$'s prediction 
		
		\STATE $x_{adv} = x_{adv} + \epsilon sign(\nabla_x J(\theta_{M_P}, x_{adv}, c))$, where
		\STATE $\theta_{M_P}$ are the parameters of $M_P$, and $J(\theta_{M_P}, x, c)$ is 
		\STATE the loss used to train $M_P$
		\eindent
		\STATE {\bfseries Return:} $x_{adv}$, return the adversarial instance
		
	\end{algorithmic}
	\label{alg:adv}
\end{algorithm}

Our generalized adversarial attack is then defined in Algorithm \ref{alg:adv}.  As input, this algorithm takes our original classifier $M_O$, an unaltered instance $x$ predicted by $M_O$ to belong to some class of interest, and a pseudo model, $M_P$, trained to predict the confidence $M_O$ will report for the class of interest.  It returns an adversarial instance, $x_{adv}$, that changes the prediction of $M_O$.  This algorithm is inspired by the fast gradient sign method \cite{Goodfellow2015}, but because $M_O$ is a black-box model we are not able to take the derivative of $x$ with respect to the weights of $M_O$.  Instead, we rely on the weights of $M_P$ to find the direction that $x_{adv}$ must move to maximize the loss and ultimately change the classifier's prediction.  More sophisticated black-box attack methods may further improve results.

To summarize, the following steps are performed to calculate the GAD value from Equation \ref{eq:GAD}: 

\begin{enumerate}
	\item Algorithm \ref{alg:extract} is used to extract a psuedo model, $M_P$, from $M_O$ (the black-box classifier being evaluated).
	\item Algorithm \ref{alg:adv} is used to create $A(M,x)$.
	\item $A(M,x)$ is used to produce an adversarial version of every instance in the evaluation dataset.  
	\item Equation \ref{eq:MAE} is used to calculate the MAE between every instance in the evaluation dataset and its adversary as found in step 3.
	\item A LOESS curve is fit to the MAE values calculated in step 4, where the independent variable is the classifier's confidence and the dependent variable is MAE, creating $F\left(\hat{p}_x\right)$.
	\item Equation \ref{eq:GAD} is used to calculate the GAD for any instance $x$
\end{enumerate}

\subsection{Search}

Once GAD is calculated for every instance in the evaluation dataset, the search for high confidence errors can proceed as described in \cite{Bennette2020}, and captured in Algorithm \ref{alg:Greedy}.  Generally, the search procedure queries the oracle to label the instances that have the lowest adversarial distance.  These instances are selected because we believe they represent predictions for which the classifier is overly confident, and therefor prone to error.  Our evidence for this belief is that the instances require less perturbation to be turned adversarial than is expected for predictions made with that confidence.  Additional search procedures that incorporate other objectives could easily be devised, such as including the spread of the data points as a penalty when selecting points via Adversarial Distance.  However, the objective of this study is to show the potential benefit of Adversarial Distance to drive a search.

\begin{algorithm}[hbt!]
	\caption{Adversarial Distance Search}
	\label{alg:Greedy}
	\begin{algorithmic}
		\STATE {\bfseries Input:} Evaluation set $\mathbb{X}$, budget $B$, and classifier $M$
		\STATE
		\STATE $Q=\{\}$, instances that have been queried
		\STATE $S = \{\}$, misclassified instances
		\STATE{\bfseries For: } $b = 1, 2, ..., B$ {\bfseries do:}
		\bindent
		\STATE $q = \argmin_{x \in \mathbb{X} \ and \ x \not\in Q} AdvDist(x)$	
		\STATE $y_{q} = OracleQuery(q)$
		\STATE $Q \leftarrow Q \cup q$
		\STATE{\bfseries If:} $y_q \neq M(q)$ {\bfseries :} $S \leftarrow S \cup q$
		\eindent
		\STATE {\bfseries Return: $Q$ and $S$}
	\end{algorithmic}
	\label{alg:search}
\end{algorithm}

\section{Results}\label{sec:Results}

This section introduces the experimental datasets, classifiers, evaluation procedure, and results.

\subsection{Datasets and Classifiers}

Details of our three experimental datasets can be found below.  Note that only datasets with numerical features were selected because it is not clear how adversarial attacks should be performed for categorical features. 

\textbf{Pang04:} The classification task for this dataset is to label sentences from IMDb summaries and Rotten Tomatoes reviews as objective or subjective \cite{pang2004sentimental}. The training dataset contains 4,000 instances, and the test dataset contains 5,000 instances.  The training dataset was intentionally  biased to contain fewer objective sentences.  The ten features used for this classification task were derived from a singular value decomposition of a bag-of-words feature set.  When searching for errors, the search was focused on finding instances misclassified as subjective.

\textbf{Phoneme:} The classification task for this dataset is to identify nasal and oral sounds using six phonemes (features) \cite{alcala2011keel}.  The training dataset contains 3,400 instances, and the test dataset contains 2,000 instances.  The training dataset was intentionally biased such that oral sounds only had positive values for the Iy phoneme.  When searching for errors, the search was focused on finding instances misclassified as oral.

\textbf{Yeast:} The classification task for this dataset is to identify types of yeast \cite{Dua2019}.  We selected two classes of yeast: NUC and MIT, due to their prevalence in the dataset. We used features V1 described as "McGeoch’s method for signal sequence recognition", and V2 described as the "score of discriminant analysis of the amino acid content of the N-terminal region (20 residues long) of mitochondrial and non-mitochondrial proteins" The Yeast dataset was split into 150 instances for training and 490 intsances for testing.  The training data was intentionally biased to select cases of MIT with high values of V1 and low values of V1 for the NUC class.  When searching for errors, the search was focused on finding instances misclassified as NUC.

A Support Vector Machine (SVM) is then used to build a classifier for each dataset using the biased training data.  The SVM training parameters are selected via a grid search over linear and radial basis function kernels.  The trained SVMs represent the black-box classifiers that we want to evaluate with some human-in-the-loop query algorithm.  Table \ref{tab:Accuracy} provides the training and test accuracies for each SVM, and the selected kernel.  The large drops between training and test accuracy are mainly attributed to the intentional biasing of the training dataset.  Drops in test accuracy were observed to be smaller when the training dataset was not biased.  

\begin{table}[hbt!]
	\begin{center}
		\begin{tabular}{@{}llllll@{}}
			\toprule
			Dataset & Train & Test & Kernel & Pseudo $R^2$ & LOESS \\ \midrule
			Pang04 &  98\% & 81\%  & Radial & 0.86 & Log MAE \\
			Phoneme &  81\% & 81\% & Radial & 0.99 & Log MAE \\
			Yeast &  92\% & 74\% & Linear & 0.99 &  MAE \\ \bottomrule
		\end{tabular}
	\end{center}
	\caption{Classifier and pseudo model performance split by dataset.  Additionally, an indicator if the LOESS line fit for calculating GAD uses MAE or Log MAE.}
	\label{tab:Accuracy}
\end{table}

\begin{figure}[hbt!]
	\begin{center}
		\includegraphics[width=1\linewidth]{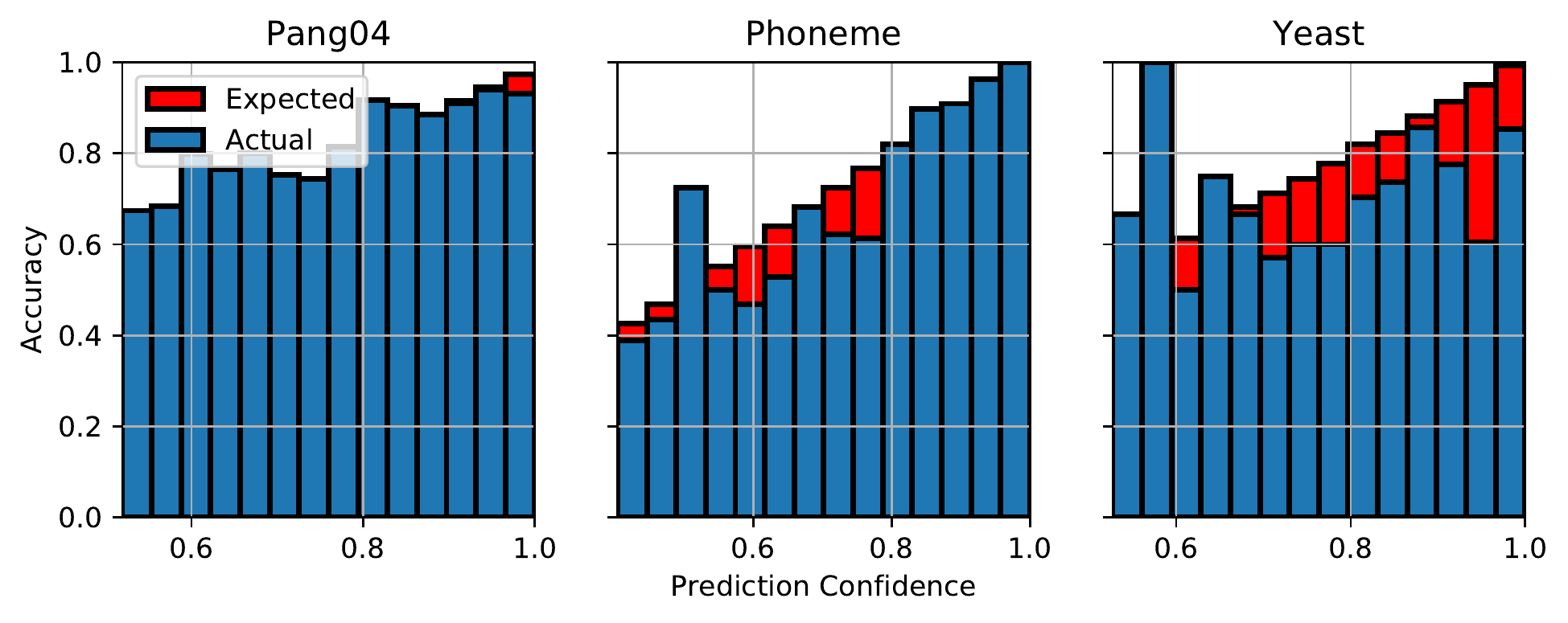}
	\end{center}
	\caption{Reliability diagram for each dataset/classifier pair. The red in each diagram indicates overconfidence. The three datasets have differing levels of overconfidence.}
	\label{fig:overconfidence}
	\label{fig:onecol}
\end{figure}

Figure \ref{fig:overconfidence} shows the overconidence profile of each test dataset and classifier pair through a reliability diagram \cite{Gao2018}.  A reliability diagram plots the observed accuracy of a classifier's predictions vs the expected accuracy of the predictions given the classifier's confidence.  Areas shown in red represent levels of classifier confidence where the model is overly confident.  The reliability diagrams in Figure \ref{fig:overconfidence} are restricted to predictions from the classes of interest, as identified in the dataset descriptions.  The reliability diagrams reveal that the three datasets represent three different overconfidence profiles.  These varying levels of overconfidence are due to the degree of the intentional biasing and the class selected to check for misclassifications.  The Pang04 dataset has very little (if any) overconfidence for the subjective class, the Phoneme dataset has overconfidence for low confidence values with the oral class, and Yeast has overconfidence for high confidence values with the NUC class.  Because the GAD search is designed to find errors due to classifier overconfidence, these varying levels of overconfidence will test the ability of our algorithm to find errors under different conditions.

Finally, a pseudo model is built for each classifier to predict the confidence the SVM classifier will have for the class of interest, given a new instance or data point.  These pseudo models are created for each classifier following the steps in Algorithm \ref{alg:extract}, where $M_O$ is the black-box classifier being evaluated (the SVM), $D$ is the unbiased test dataset, the size of the design space, $N$, is 50,000, the architecture of the extraction model, $M_E$, is a Neural Network with 5 dense layers, $e$ is 20 training epochs, and the class of interest aligns with that identified in the dataset description.  As previously discussed, a pseudo model is needed because the evaluation is being performed under the assumption that we are evaluating a black-box classifier with no access to its training data or architecture.  The black-box classifier will only provide class predictions and confidences.  Table \ref{tab:Accuracy} shows the $R^2$ value for each pseudo model.  Subjectively, these $R^2$ values indicate that the pseudo models have captured the behavior of the black box classifier.  Please refer to Section \ref{subsec:GAA} for a discussion on extracting the pseudo models.

\subsection{Evaluation}

The purpose of our evaluation is to determine if the GAD search can discover errors at a rate greater than expected given the confidence of the classifier.  Meaning, we would like to determine if the search method has taken advantage of some underlying structure of the classifier to guide users to errors, and is not simply finding errors through random chance.  In the case of the GAD search, we attempt to provide users with predictions for which the classifier is overly confident in its prediction.  As previously motivated in \cite{Maurer2018} and \cite{Bennette2020}, we will measure the value of our query sets through SDR which is defined in Equation \ref{eq:sdr}.  Briefly, SDR can be viewed as the ratio of discovered errors to the expected number of errors given the classifier's confidence for the queried points.  We also record the number of errors discovered, and the confidences of the points sampled.

\subsection{Experiments}

Our proposed GAD search is compared to the Lakkaraju search \cite{Lakkaraju2016}, the Bansal and Weld search \cite{Bansal2018}, the Facilities Location search \cite{Maurer2018}, and to a random search. To encourage follow on work, all of the code used to perform our experiments will be available at https://github.com/afrl-ri/adversarialDistance. Code for the Lakkaraju and Bansal and Weld search was made available in \cite{Bansal2018}, and the Facilities Location search code was made available in \cite{Maurer2018}.  Both were used for this experimentation. Results analysis was done with R and the tidyverse packages \cite{r2017} \cite{tidy2017}.

Because of the sensitivity of the search techniques to initial conditions, each search is run 100 times using a random 250 instance subset of the test data. This replication strategy simulates 100 unlabeled evaluation datasets for each classifier and search method. Each evaluation set only contains instances predicted by the classifier to belong to the class of interest (defined in the dataset description) with confidence greater than 65\% (the threshold used in previous works to denote a high-confidence error). Each search selects a 50 sample query set and is compared using SDR, the number of discovered errors, and the confidence of the queried points.  We use the ground truth labels of the simulated evaluation set to verify the correctness of predictions instead of a human-in-the-loop.

\begin{figure}[hbt!]
	\begin{center}
		\includegraphics[width=0.9\linewidth]{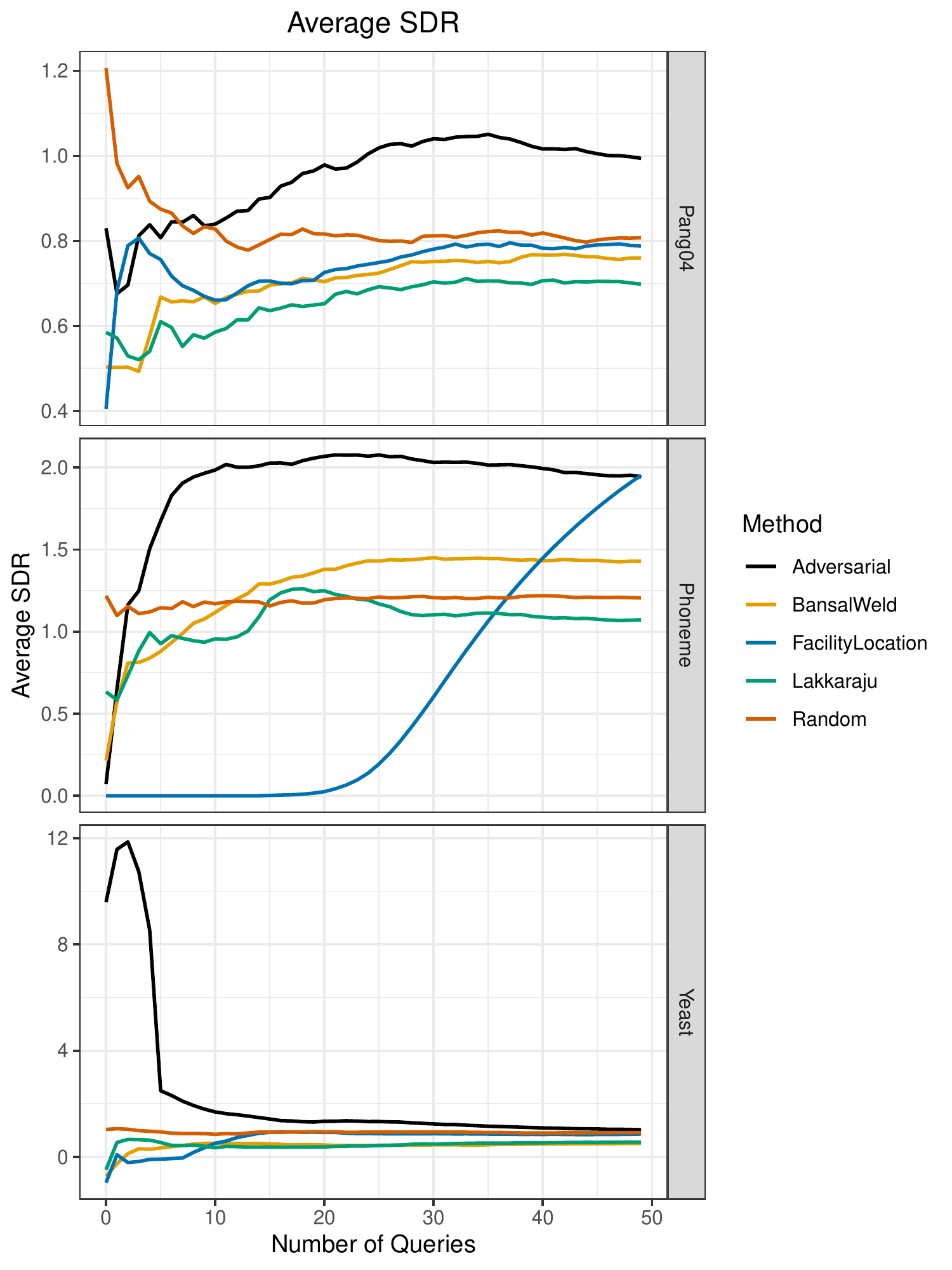}
	\end{center}
	\caption{Average SDR across 100 runs of the search methods. Recall that an SDR of one indicates that errors are discovered at the rate expected given the confidence of the sampled points.  \textbf{Note: a log scale is used for the SDR values with the Yeast dataset.}}
	\label{fig:sdr}
\end{figure}

\begin{figure}[hbt!]
	\begin{center}
		\includegraphics[width=0.9\linewidth]{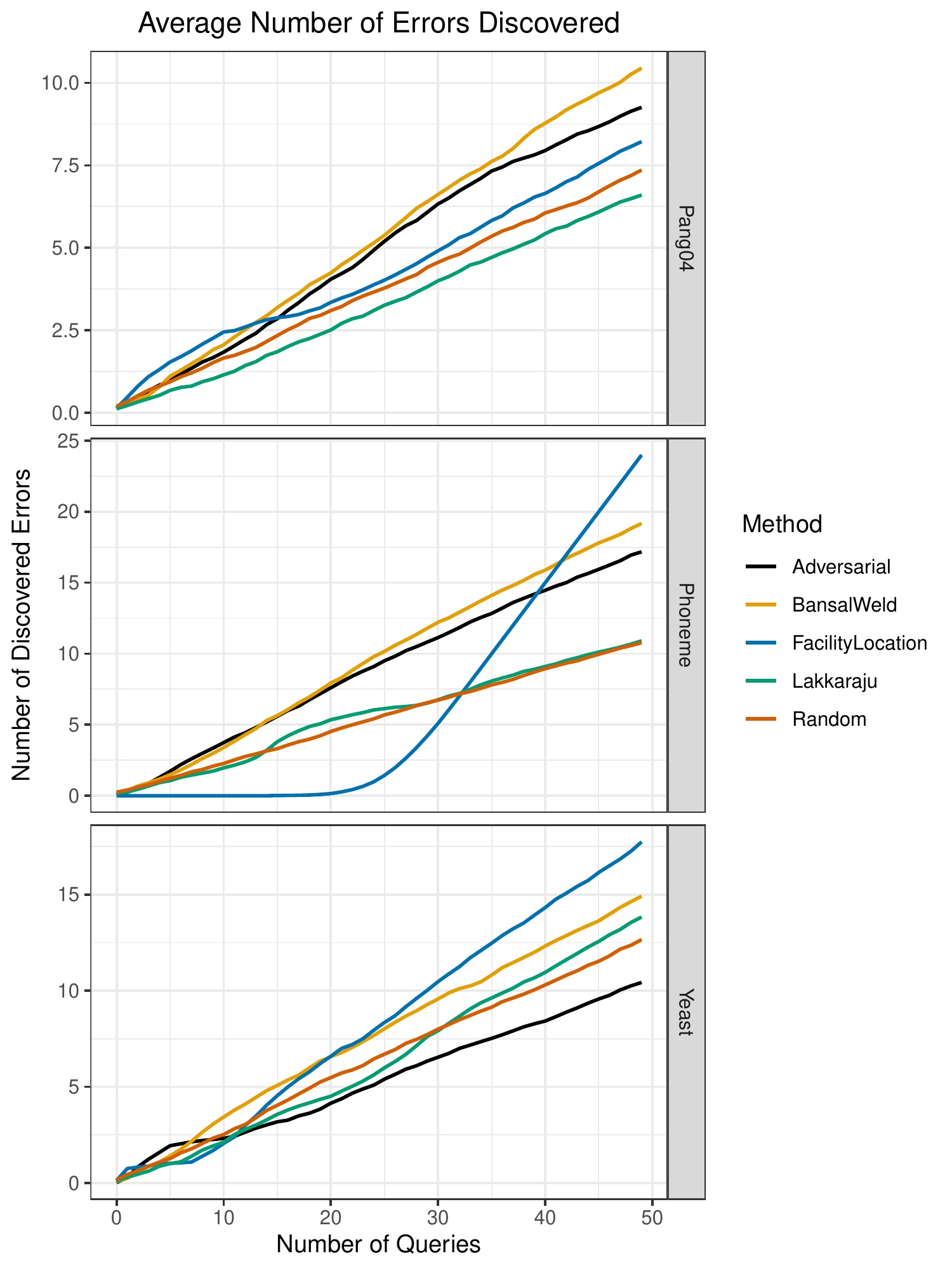}
	\end{center}
	\caption{Mean number of errors discovered  across 100 runs of the search methods.  Although the Adversarial Search does not find the most errors, it finds errors at a rate greater than expected given the confidence of the sampled points (as indicated by SDR values greater than one)}
	\label{fig:fix}
\end{figure}

\begin{figure}[hbt!]
	\begin{center}
		\includegraphics[width=0.9\linewidth]{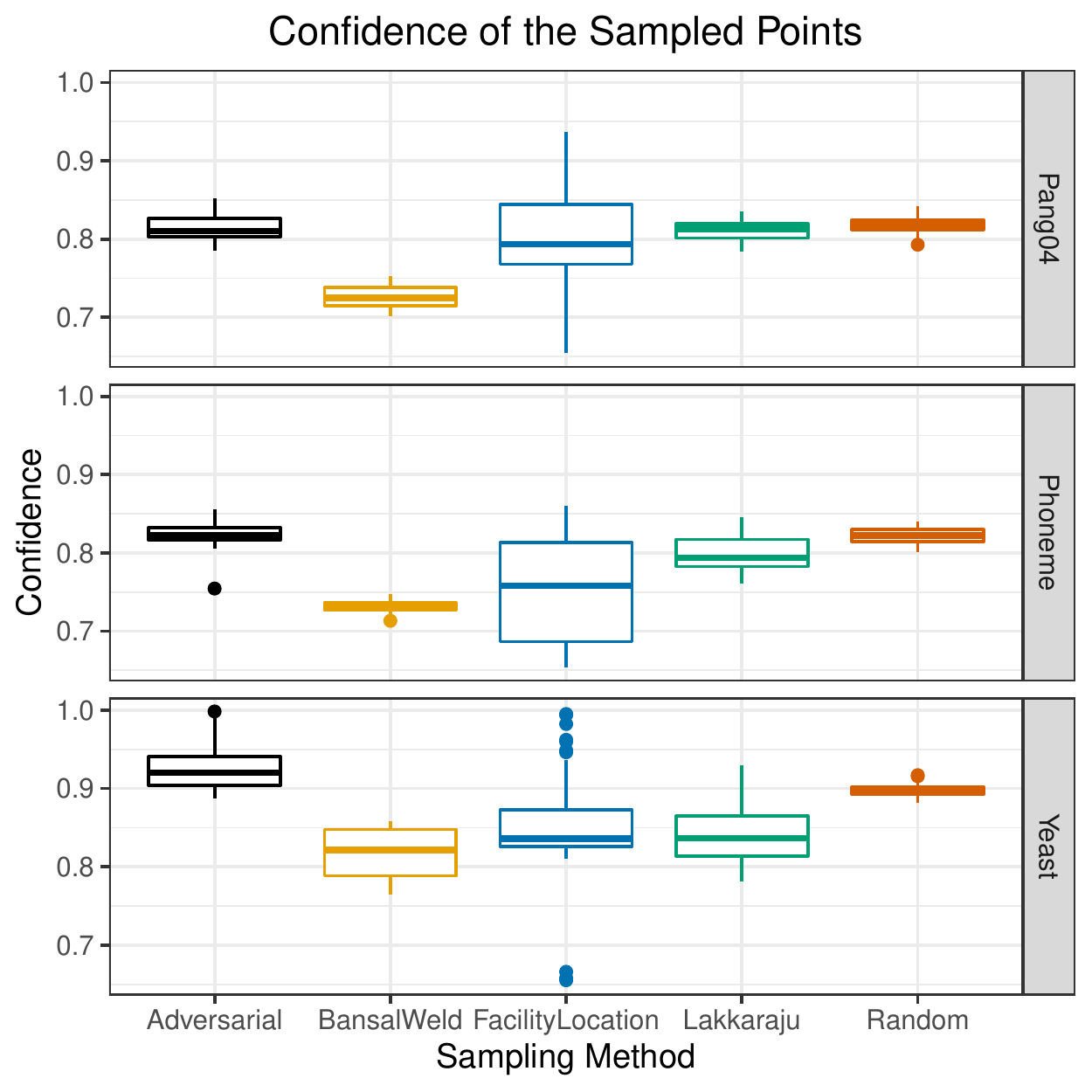}
	\end{center}
	\caption{Box plot showing the model confidence of the sampled data points.}
	\label{fig:con}
\end{figure}

Figure \ref{fig:sdr} shows the average SDR for each dataset and classifier pair for the compared methods.  In all cases, the GAD search dominates the curves of the competing methods.  Note that the SDR of the Yeast dataset is presented in a log scale because there is a large amount of overconfidence that the methods can discover, which has led to large SDR values.  This transformation allows the behavior of the methods to still be inspected.  Figure \ref{fig:fix} shows the number of errors discovered for each dataset and method.  Although the GAD search does not always find the most errors, it finds a competitive number of errors, and more importantly, it generally finds more errors than should be expected given the confidence of the classifier (indicated by an SDR greater than 1).  Finally, Figure \ref{fig:con} shows the confidence of the points queried by each search method.

Recall the reliability diagram of Figure \ref{fig:overconfidence} when considering the Pang04 dataset in Figure \ref{fig:sdr}.  There is very little, if any, overconfidence for the search methods to discover when considering the subjective class.  As a result, the GAD search (which is designed to search for overconfidence) achieves an SDR very near one.  Meaning, the amount of discovered error is not surprising.  However, all other search methods find fewer errors than should be expected, indicated by SDRs less than one.  The GAD search seems to be revealing that the Pang04 classifier is well behaved for the subjective class, while the other search methods return even less helpful results.

When considering the Phoneme dataset, it is interesting to see that the GAD search quickly achieves an SDR near two, and maintains this value.  This indicates that the GAD search helps find twice as many errors as should be expected given the confidence of the classifier for the queried points.  Other than the Facility Location search, the remaining methods hover around an SDR of one, or just above.  Interestingly, after a sufficient number of samples the Facility Location search also achieves an SDR of two.  This is likely because the Facility Location search builds a meta-model to drive its selection procedure.  As indicated in Figure \ref{fig:fix}, it typically takes some time for the Facility Location search to find an error from which it can start to build its meta-model.  Once an error is discovered, it seems the meta-model quickly identifies a pattern of the errors and achieves an SDR greater than one.  Due to the slow start of the facility location search, we believe it may be useful to prime its meta-model with errors discovered by the GAD search technique.

Finally, inspecting the SDR results of the Yeast dataset in Figure \ref{fig:sdr}, it shows that all of the search techniques achieve an SDR greater than one (recall results are displayed on a log scale for this dataset).  This is not surprising because of the amount of overconfidence shown in Figure \ref{fig:overconfidence}.  However, it is clear that the GAD search quickly finds errors at a rate greater than expected, and continues to do so at a rate greater than the competing methods.  However, the SDR of the GAD search decreases with the number of queried points, providing further evidence that it may be beneficial to prime a meta-model search technique such as the Facility Location search with points initially queried by the GAD search.

\section{Conclusions}\label{sec:Conclusions}

In this work we proposed a generalization of Adversarial Distance to help a human-in-the-loop search  discover high-confidence errors made by a classification model for an unlabeled evaluation dataset.  Adversarial Distance is a measure of how much perturbation must be applied to an instance for the targted classifier to change its prediction, compared to how much perturbation should be expected given the confidence of the classifier's prediction.  Previous works defined Adversarial Distance exclusively for image classification tasks, but here we extend the measure by proposing a generalized adversarial attack sufficient for cacluating Adversarial Distance.  

Through experimental evaluation, our generalized measure is shown to direct users towards high-confidence errors at rates greater than should be expected given the confidence of the black-box classifier being evaluated.  This means that the Generalized Adversarial Distance reveals something informative about the error structure of the model, and is not finding errors by simple chance.  Additionally, results show the Generalized Adversarial Distance outperforms competing methods previously proposed in the literature.

\subsection{Opportunities for Improvement}

This work has shown that Adversarial Distance is a promising tool to drive searches for high-confidence errors.  However, the proposed generalization strategy relies on an adversarial attack that leverages information from a pseudo model to create adversarial instances.  Although the proposed algorithm for pseudo model extraction worked well for the experimental datasets in our study, we believe it will suffer when the dimensionality of the data increases.  In those cases it may be beneficial to adjust the model extraction algorithm, or to use a black-box attack specifically designed for your data modality.

Additionally, we observe that the Generalized Adversarial Distance was very effective at quickly finding high-confidence errors.  However, as the number of discovered errors increases, the rate of discovery decreases or stays the same (as observed through values of SDR).  Therefore, we believe it may be beneficial to use the errors initially found from the Generalized Adversarial Distance search to prime a search that utilizes a meta-model to learn about the error structure of the model being evaluated.  For example, the Facility Location search achieves high values of SDR once it discovers a sufficient number of errors, but it can take a large number of queries for the search to find its first error (as in the Yeast dataset).

Finally, reviewers wondered how effective the proposed method would be at finding high confidence errors when evaluating classifiers that defend against adversarial attacks.  Unfortunately, there was not enough time to run this analysis before final submission, but we believe this could lead to a very interesting study and help improve our methodology.

\bibliographystyle{IEEEtran}
\bibliography{library}

\end{document}